# Optimal path planning and weighted control of a four-arm robot in on-orbit servicing

Celia Redondo-Verdú
*Department of Physics, Systems Engineering and Signal Theory*
*University of Alicante*
*Alicante, Spain*
celia.redondo@ua.es

José L. Ramón
*Department of Physics, Systems Engineering and Signal Theory*
*University of Alicante*
*Alicante, Spain*
jl.ramon@ua.es

Álvaro Belmonte-Baeza
*Department of Computer Sciences and Artificial Intelligence*
*University of Alicante*
*Alicante, Spain*
alvaro.belmonte@ua.es

Jorge Pomares*
*Department of Physics, Systems Engineering and Signal Theory*
*University of Alicante*
*Alicante, Spain*
jpomares@ua.es
*Correspondig author

Leonard Felicetti
*School of Aerospace, Transport and Manufacturing*
*Cranfield University*
*Cranfield, UK*
Leonard.Felicetti@cranfield.ac.uk

*Abstract*— This paper presents a trajectory optimization and control approach for the guidance of an orbital four-arm robot in extravehicular activities. The robot operates near the target spacecraft, enabling its arm's end-effectors to reach the spacecraft's surface. Connections to the target spacecraft can be established by the arms through specific footholds (docking devices). The trajectory optimization allows the robot path planning by computing the docking positions on the target spacecraft surface, along with their timing, the arm trajectories, the six degrees of freedom body motion, and the necessary contact forces during docking. In addition, the paper introduces a controller designed to track the planned trajectories derived from the solution of the nonlinear programming problem. A weighted controller formulated as a convex optimization problem is proposed. The controller is defined as the optimization of an objective function that allows the system to perform a set of tasks simultaneously. Simulation results show the application of the trajectory optimization and control approaches to an on-orbit servicing scenario.

## I. Introduction

Robotic servicing spacecraft may undertake a broader range of tasks, including complex operations that require increased levels of autonomy, reliability, efficiency, and safety. The use of several robotic arms in on-orbit servicing operations were proposed in missions such as NASA's On-orbit Servicing, Assembly and Manufacturing – 1 (OSAM-1) [1] and the DARPA's Robotic Servicing of Geosynchronous Satellites (RSGS) [2]. Robots will play a vital role in tasks such as constructing habitats for humans on Mars or the lunar surface [3], as well as building telescopes and large-scale space structures [4]. Space robotics tasks such as on-orbit manipulation, in-situ resource utilization or on-orbit servicing will require complex manipulation systems composed of multiple arms.

This paper presents a trajectory optimization and control approach for the guidance of a multi-arm robot in extravehicular activities. More specifically, a multi-arm robot with four arms is considered. The robot operates near the target spacecraft, enabling its arm's end-effectors to reach the spacecraft's surface. Connections to the target spacecraft can be established by the arms through specific footholds (docking devices) or by gripping onto structural elements. The future application of the robot in on-orbit manufacturing tasks justifies the number of arms considered for the robot. The proposed trajectory optimization approach automatically computes the robot body motion and the arms trajectories and interaction forces required to guide the robot towards a given location by using the docking devices.

Space robots work within environments that are inherently unpredictable and subject to various disturbances. Generally, these robots have redundant degrees of freedom, enabling them to navigate through uncertain workspaces and perform multiple tasks simultaneously or sequentially. The use of intelligent motion planning and control systems allow the robots to work and to adapt their performance to non-deterministic workspaces with uncertainties. Optimization-based trajectory planning and control systems are recently used in the literature for the guidance and control of space robotics. These approaches have been proposed for trajectory optimization of spacecraft [5] and for trajectory planning of free-floating space-robotic systems [6], where convex programming has been employed to identify locally feasible kinematic paths. New optimization-based motion planning and control methods for space robotics have been proposed in the last few years for complex on-orbit manipulation systems. In [7], a strategy for optimal motion planning and control is outlined, aimed at facilitating the

This research was supported by the grant CIAICO/2022/077 "Plataforma de control y simulación de código abierto para escenarios de robótica de servicio en órbita" (Programa AICO 2023, Conselleria de Innovación, Universidades, Ciencia y Sociedad Digital de la Generalitat Valenciana, Spain), and by the Spanish Ministry of Universities under grant FPU21/02586.

dissipation of momentum from a tumbling satellite held by a space manipulator. Furthermore, [8] describes the motion planning of a dual-arm space robot manipulator subjected to external forces and moments while operating in space. In [9], a trajectory optimization method is proposed for the guidance of a two-arm humanoid robot performing on-orbit manipulation tasks by using a simplified robot dynamics based on single rigid body dynamics. In the last years, trajectory optimization approaches have been applied to control multilegged robots on earth [10][11]. In these approaches, the continuous-time optimization task is expressed by using a set of constraints and decision variables [12]. In this paper, this approach is extended to the specific problem of the guidance of the on-orbit four-arm robot. Therefore, additional constraints have been introduced to take into account the robot specific dynamics and perturbations. A non-linear programming solver is employed to create complex trajectories, enabling the robot to reach a specified destination by coordinating its arm motions and the docking devices present on the target spacecraft.

The readiness level of space-certified cameras and onboard computers has advanced sufficiently to enable the adaptation of ground-based robot visual servoing techniques for onboard implementation. For example, in [13] the relative position and orientation of a spacecraft is obtained with respect to the International Space Station through the identification of visual features on a designated target. We should mention previous works that propose the application of direct visual servoing systems to solve problems involving the guidance of multiple spacecraft [14], on-orbit manipulation [15], or redundant manipulators in space [16]. In this paper, the proposed trajectory optimization method uses the visual feedback provided by a camera located at the robot body. From this visual information, a map of the target spacecraft is obtained that is used to generate the robot arms and body path planning.

The trajectory optimization process involves determining the docking points on the target spacecraft surface, along with their timing, the arm movements (taking into account the robot's kinematics and dynamics), the robot body motion, and the required interaction forces during docking. However, a control approach should be designed to track the trajectories derived from the solution of the nonlinear programming problem. This paper proposes a weighted controller formulated as a convex optimization problem. To solve the robot redundancy, the controller is defined as the optimization of an objective function that allows the system to perform a set of tasks simultaneously. The robot specific dynamics and task constraints are included. As an example, the controller performs two tasks: tracking joint trajectories for each arm and tracking the robot body trajectory. In order to highlight the controller performance, the results section shows a comparison of the proposed controller with respect other previous defined for the guidance of on-orbit robots.

The remaining part of the paper is divided into the following sections. Section 2 describes the system architecture, on-orbit servicing scenario, and the robot kinematics and dynamics. The proposed trajectory optimization problem is presented in Section 3. Section 4 describes the multi-arm trajectory control method. Simulation results are detailed in Section 5, and main conclusions are presented in Section 6.

## II. SYSTEM ARCHITECTURE AND DYNAMICS

In this paper, the path planning and control system of an on-orbit multi-arm robot is proposed. The on-orbit servicing scenario is represented in Fig. 1, where the robot and the target spacecraft surface are represented. The robot has $\zeta$ arms with $n$ degrees of freedom each. In the results section, a robot with $\zeta = 4$ arms a $n = 6$ degrees of freedom each is considered (see Fig. 1). The robot arms have a docking system at their end-effectors. A set of docking devices are located on the target spacecraft surface. In Fig. 1 a planar target spacecraft surface is considered, where each docking device is positioned with a 0.5 meter spacing between them. The joint coordinates of each arm are represented as $q_i \in \Re^n$ ($i = 1 \ldots \zeta$). The coordinate frame $B$ of the robot is situated at the center of its body. The robot aims to be in close proximity to the target spacecraft surface, enabling its arm end-effectors to reach the docking points on the surface of the target spacecraft. In addition, a three-dimensional map of the workspace or target spacecraft surface is created from the point cloud captured by a camera located in the robot body.

The system dynamics shows the relationship between the acceleration, forces, and torques arising at the base of the robot and each arm. Specifically, the system dynamics relate to the linear and angular accelerations of the robot base $\dot{v}_b = [\ddot{t}_b^T, \dot{\omega}_b^T]^T \in \Re^6$ with respect to the inertial coordinate frame, the joint accelerations of each arm, $\ddot{q}_i$, the forces and torques exerted on the robot base, $h_b \in \Re^6$, and the torques applied on each arms' joint, $\tau \in \Re^n$. This relationship can be defined as:

$$\begin{bmatrix} h_b \\ \tau_i \end{bmatrix} = \begin{bmatrix} M_{bb} & M_{bi} \\ M_{bi}^T & M_{ii} \end{bmatrix} \begin{bmatrix} \dot{v}_b \\ \ddot{q}_i \end{bmatrix} + \begin{bmatrix} c_b \\ c_{mi} \end{bmatrix} \quad (1)$$

where $M_{bb} \in \Re^{6\times 6}$ is the inertia matrix of the robot base, $M_{bi} \in \Re^{6\times n}$ is the coupled inertia matrix of the base and the arm $i$, $M_{ii} \in \Re^{n\times n}$ is the inertia matrix of the arm $i$, and $c_b$, and $c_{mi}$, $\in \Re^6$ are a velocity/displacement-dependent, non-linear terms for the base spacecraft and arm $i$, respectively.

Equation (1) can be extended by including the environmental perturbation torques and rewritten in the following form:

$$M_i^* \ddot{q}_i + C_i^* = \tau_i + \tau_{envi} \quad (2)$$

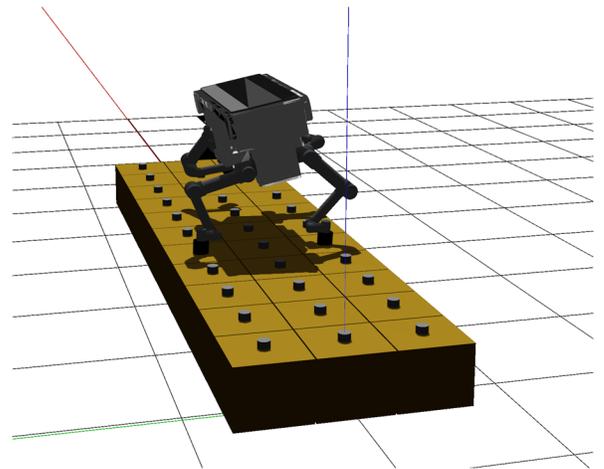

Fig. 1. On-orbit servicing scenario

where $\boldsymbol{\tau}_{envi} \in \Re^n$ are the environmental perturbation torques acting on arm $i$, $\boldsymbol{M}_i^* \in \Re^{n \times n}$ is the generalised inertia matrix, and $\boldsymbol{C}_i^* \in \Re^{ne}$ is the generalised Coriolis and centrifugal vector for the arm $i$, defined explicitly as:

$$\boldsymbol{M}_i^* = \boldsymbol{M}_{ii} - \boldsymbol{M}_{bi}^T \boldsymbol{M}_{bb}^{-1} \boldsymbol{M}_{bi} \quad (3)$$

$$\boldsymbol{C}_i^* = \boldsymbol{c}_{mi} - \boldsymbol{M}_{bi}^T \boldsymbol{M}_{bb}^{-1} \boldsymbol{c}_b \quad (4)$$

For a given arm $i$, the linear and angular momenta of the system $[\boldsymbol{\ell}^T, \Psi^T]_i^T \in \Re^6$ are defined as:

$$\begin{bmatrix} \boldsymbol{\ell} \\ \boldsymbol{\Psi} \end{bmatrix}_i = \boldsymbol{M}_{bb} \boldsymbol{v}_b + \boldsymbol{M}_{bi} \dot{\boldsymbol{q}}_i \quad (5)$$

where $\boldsymbol{v}_b = [\dot{\boldsymbol{t}}_b^T, \boldsymbol{\omega}_b^T]^T \in \Re^6$ is the linear and angular velocities of the robot base with respect to the inertial coordinate frame, and $\dot{\boldsymbol{q}}_i \in \Re^n$ is the time derivative of the joint vector of the arm $i$.

The arm Jacobian, $\boldsymbol{J}_i \in \Re^{6 \times n}$, provides the following relationship:

$$\dot{\boldsymbol{p}}_i = \boldsymbol{J}_i \dot{\boldsymbol{q}}_i + \boldsymbol{J}_b \boldsymbol{v}_b \quad (6)$$

where $\dot{\boldsymbol{p}}_i \in \Re^6$ is the linear and angular velocity of the arm end-effector in the inertial frame, and $\boldsymbol{J}_b \in \Re^{6 \times 6}$ is the Jacobian matrix of the robot. Combining (6) with (5) yields an equation that directly relates the joint speeds and end-effector motion of the arm $i$:

$$\dot{\boldsymbol{p}}_i = \boldsymbol{J}_{gi} \dot{\boldsymbol{q}}_i + \boldsymbol{J}_b \boldsymbol{M}_{bb}^{-1} \begin{bmatrix} \boldsymbol{\ell} \\ \boldsymbol{\Psi} \end{bmatrix}_i \quad (7)$$

$$\boldsymbol{J}_{gi} = \boldsymbol{J}_i - \boldsymbol{J}_b \boldsymbol{M}_{bb}^{-1} \boldsymbol{M}_{bi} \quad (8)$$

where $\boldsymbol{J}_{gi}$ is the Generalised Jacobian Matrix for the arm $i$.

## III. TRAJECTORY OPTIMIZATION

This section describes the trajectory optimization approach for the path planning of the robot in the on-orbit servicing application described in Section 2. The objective of this method is to determine the arms and body motion to guide the robot from an initial location $[\boldsymbol{t}_b^T(t=0) = \boldsymbol{t}_{b0}^T, \boldsymbol{\phi}_b^T(t=0) = \boldsymbol{\phi}_{b0}^T]^T$ towards a given desired location $[\boldsymbol{t}_b^T(t=T) = \boldsymbol{t}_{bd}^T, \boldsymbol{\phi}_b^T(t=T) = \boldsymbol{\phi}_{bd}^T]^T$, considering $T$ as the desired duration of the maneuver. The algorithm can schedule and plan the movements of the $\zeta$ arms to move the robot by using the docking devices present outside the target spacecraft. Therefore, the trajectory optimization method generates the trajectory for the robot base $[\boldsymbol{t}_b^T(t), \boldsymbol{\phi}_b^T(t)]^T$, and the interaction forces and end-effector trajectories for each arm $i$, $\boldsymbol{f}_i(t)$, and $\boldsymbol{p}_i(t)$ respectively. The interaction forces and end-effector trajectories for each arm are automatically generated by the proposed method by considering that each arm has two kinds of phases: docking phases ($t \in \mathcal{C}_i$) and non-docking phases ($t \notin \mathcal{C}_i$). Therefore, for a given arm $i$, the duration of these phases is denoted as $\Delta T_{ci,j}$ and $\Delta T_{nci,j}$ respectively, where $j = 1 \dots N$, and $N$ being the number of docking phases for each arm. Hence, the maneuver duration should be $\sum_{j=1}^{N} \Delta T_{ci,j} + \Delta T_{nci,j} = T$, and the planned trajectory and interaction forces are $\boldsymbol{p}_i(t, \Delta T_{ci,1}, \Delta T_{nci,1} \dots)$, and $\boldsymbol{f}_i(t, \Delta T_{ci,1}, \Delta T_{nci,1} \dots)$, respectively. These trajectories are codified by using different polynomials of fixed durations that are joined to create a continuous spline whose coefficients are optimized. For example, for a given arm's trajectory, $\boldsymbol{p}_i(t \notin \mathcal{C}_i)$, multiple third-order polynomials are considered per non-docking phase, and a constant value for the docking phases. For each arm's force profile, $\boldsymbol{f}_i(t \in \mathcal{C}_i)$, multiple polynomials represent each docking phase and zero force is stablished during the non-docking phase. The duration of each phase, and with that the duration of each arm's polynomial, is changed based on the optimized phase durations $\Delta T_{ci,j}, \Delta T_{nci,j}$.

Various constraints are also defined to ensure realistic motions by guaranteeing the robot kinematic and dynamic properties. The kinematic constraint guarantees that the range of motions of each arm, $i$, $\mathcal{K}_i$ is consistent with the robot kinematic. The kinematics constraint is defined as a prism, with edge $2\xi_i$, centered at nominal position for each arm, $i$:

$$|\boldsymbol{R}_b[\boldsymbol{p}_i(t) - \boldsymbol{t}_b^T(t)] - \boldsymbol{p}_{ni}| < \xi_i \quad (9)$$

where $\boldsymbol{p}_{ni}$ is the nominal position for each arm end-effector, and $\boldsymbol{R}_b$ represents the attitude of the robot base with respect to the inertial frame (rotation matrix). This kinematic constraint allows to guarantee the motion range for each arm and simultaneously avoiding self-collisions. Once the range of motion is guaranteed, the dynamic constraint should also be considered in order to obtain realistic motions. To do this, Equation (2) is included as a system constraint.

Additional constraints should be included in the docking phase, $t \in \mathcal{C}_i$, to guarantee the docking positions are achieved. These constraints should be applied during the docking phases, $\Delta T_{ci,j}$. The set of docking positions are defined by the vector $\varphi$. Therefore, in a docking phase, $\boldsymbol{p}_{ic}(t \in \mathcal{C}_i) \in \varphi$. In addition, the $z$ coordinate of this docking point should be in the 3D map, $m(x, y)$, obtained by the camera held by the robot, i.e., $p_{ic}^z(t \in \mathcal{C}_i) = m(p_i^x, p_i^y)$. This 3D map gives information of the depths or $z$ component from the surface $x, y$ coordinates with respect to the coordinate frame $B$. In addition, the arm end-effector shouldn't slip during the docking phase, and the following constraint is included, $\dot{\boldsymbol{p}}_i((t \in \mathcal{C}_i)) = \boldsymbol{0}$, to guarantee that the docking position is kept, $\boldsymbol{p}_i(t \in \mathcal{C}_i) = \boldsymbol{p}_{ic}$.

No contact forces should be generated when $t \notin \mathcal{C}_i$, i. e., the arm is not in a docking phase. Therefore, in these phases, the following constraint has been included: $\boldsymbol{f}_i(t \notin \mathcal{C}_i) = \boldsymbol{0}$. Additional constraints are included in order to generate realistic motions for the robot. These constraints are summarized, Table 1.

An additional constraint is included to avoid collisions with the environment. To do this, the robot body must maintain a given distance, $\delta$, with respect the target surface. Therefore, the following constraint has been included:

$$p_{ic}^z(t \in \mathcal{C}_i) - m(p_{ic}^x, p_{ic}^y) > \delta \quad (10)$$

Finally, in order to minimize the contact forces during the task, the following cost function is minimized during the optimization:

$$\int_0^T \sum_{i=1}^{\zeta} \left[ \sigma_{i1} \left(f_i^x(t)\right)^2 + \sigma_{i2} \left(f_i^y(t)\right)^2 + \sigma_{i3} \left(f_i^z(t)\right)^2 \right] dt \tag{11}$$

where $f_i^x(t)$, $f_i^y(t)$, and $f_i^z(t)$ are the end-effector interaction forces in $x$, $y$, and $z$ directions, and the weights $\sigma_i$ are constant values.

TABLE I. SYSTEM CONSTRAINTS

| Constraint | Value |
|---|---|
| Initial robot location | $t_b^T(t=0) = t_{b0}^T, \phi_b^T(t=0) = \phi_{b0}^T$ |
| Final robot location | $t_b^T(t=T) = t_{bd}^T, \phi_b^T(t=T) = \phi_{bd}^T$ |
| Maneuver duration | $\sum_{j=1}^{N} \Delta T_{ci,j} + \Delta T_{nci,j} = T$ |
| Dynamics, kinematics | Equations (2), (9) |
| When $t \in \mathcal{C}_i$ | |
| End-effector doesn't slip | $\dot{p}_i(t \in \mathcal{C}_i) = 0$ |
| End-effector in docking | $p_{ic}(t \in \mathcal{C}_i) \in \varphi, p_i(t \in \mathcal{C}_i) = p_{ic}$ |
| Safety distance | $p_{ic}^z(t \in \mathcal{C}_i) - m(p_{ic}^x, p_{ic}^y) > \delta$ |
| When $t \notin \mathcal{C}_i$ | |
| No interaction forces | $f_i(t \notin \mathcal{C}_i) = 0$ |

## IV. ROBOT TRAJECTORY CONTROL

This section describes the control approach defined for tracking the trajectories obtained from the trajectory optimization approach presented in the previous sections. These trajectories cannot be reproduced in open loop by the robot because it is important to consider the robot's real state and free-floating dynamics and perturbations. The trajectory optimization trajectories are denoted in this section as the desired trajectories or tracked trajectories, $p_{di}(t)$, and the corresponding joint trajectories for each arm are represented as $q_{di}(t)$. A weighted controller defined as a convex optimization problem is proposed.

### A. Control objective

The controller objective is to determine the values of the accelerations of all the robot degrees of freedom, end-effector forces and forces/torques to be applied in all the actuated joints denoted respectively as $\ddot{q}$, $f$ and $\tau$. To obtain these control variables, the controller is defined as the optimization of an objective function that allows the system to perform a set of tasks. This set of $m$ tasks is defined as $\psi_k, k = 1 \ldots m$. As an example, in this paper a set of $m = 2$ tasks are considered. The first task allows the tracking of the desired trajectories in the joint space ($q_{di}, \dot{q}_{di}, \ddot{q}_{di}$). In this case, the system feedback is the current joint position and velocities of robot's arms ($q_i, \dot{q}_i$). The second task guarantees the tracking of the robot's body position and attitude. In this second task, the desired state or input are the body desired position $t_{bd}^T$, and the desired attitude $\phi_{bd}^T$, and its derivatives obtained from the proposed path planning algorithm. These tasks will be described in the next paragraphs in greater detail.

The tasks optimization is defined as:

$$\min_{\ddot{q}, f, \tau} \sum_{k=1}^{m} \frac{1}{2} \rho_k \|\ddot{\psi}_{rk}\|^2 \tag{12}$$

where $\psi_{rk}$, are the desired tasks, and $\rho_k$ is the weight for each task. Each task $k$, $\psi_k$, is defined as the error between the setpoint, $c_{dk}$, and the corresponding magnitude obtained from the robot joint configuration, $m_k(q)$:

$$\psi_k = c_{dk} - m_k(q) \tag{13}$$

The specific value for $c_{dk}$ and $m_k(q)$ depend on the task as will be described in the next subsection. A second order lineal dynamics is imposed on the task with the objective of obtaining the task to be minimized:

$$\ddot{\psi}_{rk} = \ddot{\psi}_k + K_d \dot{\psi}_k + K_p \psi_k \tag{14}$$

where $K_p$ and $K_d$ are proportional and derivative positive definite matrices, respectively. Therefore, the minimization of each one of the desired tasks can be defined as:

$$\min_{\ddot{q}, f, \tau} \|\ddot{\psi}_{rk}\|^2 = \min_{\ddot{q}, f, \tau} \|\ddot{\psi}_k + K_d \dot{\psi}_k + K_p \psi_k\|^2 \tag{15}$$

In the next equations, the first and second derivative of the task definition given in (13) is obtained:

$$\dot{\psi}_k = \dot{c}_{dk} - \frac{\partial m_k(q)}{\partial q} \frac{\partial q}{\partial t} = \dot{c}_{dk} - J_{\psi k} \dot{q} \tag{16}$$

$$\ddot{\psi}_k = \ddot{c}_{dk} - \dot{J}_{\psi k} \dot{q} - J_{\psi k} \ddot{q} \tag{17}$$

From (15) and (17) the task optimization can be defined as:

$$\min_{\ddot{q}, f, \tau} \|\ddot{\psi}_{rk}\|^2 = \min_{\ddot{q}, f, \tau} \|\ddot{c}_{dk} - \dot{J}_{\psi k} \dot{q} - J_{\psi k} \ddot{q} + K_d \dot{\psi}_k + K_p \psi_k\|^2 \tag{18}$$

Two kinds of trajectories to optimize are considered: tracking the joint trajectories of the robot arms, and tracking the position and attitude of the robot body. In the first case, it is necessary to compute the task Jacobian, $J_{\psi k}$, and the corresponding derivative $\dot{J}_{\psi k}$, to allow the tracking of the desired joint trajectory ($q_{di}, \dot{q}_{di}, \ddot{q}_{di}$) from the joint feedback ($q_i, \dot{q}_i$). This task Jacobian is a diagonal positive semidefinite matrix. In this case, from (13) and (16), $m_k(q) = q$, and $\frac{\partial m_k(q)}{\partial q} = 1$. Therefore, each value of the task Jacobian $J_{\psi k_{v,v}} = 1$ when the joint $v$ can be actuated and 0 if the corresponding joint is unactuated. With respect to the Jacobian derivative $\dot{J}_{\psi k}$, we can see that $\frac{\partial^2 m_k(q)}{\partial q^2} = 0$ as the values of the Jacobian are constant. In a similar way it can be computed the task for the tracking of the robot body center of mass. In this case, $J_{\psi k}$ corresponds with the Jacobian of the center of mass.

### B. System constraints

Equation (2) represents the system dynamics that should be verified to accomplish realistic trajectories for a given arm $i$. A matrix representation of the system dynamics is considered so the optimizer can obtain the solution for the variables $\ddot{q}_i, f_i, \tau_i$:

$$[M_i^* \quad -J_i^T \quad -I] \begin{bmatrix} \ddot{q}_i \\ f_i \\ \tau_i \end{bmatrix} = -C_i^* + \tau_{envi} \tag{19}$$

where $f_i$ are the end-effector interaction forces applied by the arm $i$.

The kinematics of the robot contacts must be static, i.e., the docking points cannot be moved with respect to the inertia reference frame. To assure this constraint from a kinematics point of view, the docking point velocity and acceleration should be zero:

$$\dot{p}_i(t \in C_i) = 0 \tag{20}$$

$$\ddot{p}_i(t \in C_i) = J_{gi}\ddot{q}_i + \dot{J}_{gi}\dot{q}_i + \dot{v}_{gmi} = 0 \tag{21}$$

where $v_{gmi} = J_b M_{bb}^{-1} \begin{bmatrix} \ell \\ \psi \end{bmatrix}_i$. Constraint (21) can be defined with respect to the variables to be optimized in a matrix form by using the following expression:

$$[J_{gi} \quad 0 \quad 0] \begin{bmatrix} \ddot{q}_i \\ f_i \\ \tau_i \end{bmatrix} = -\dot{J}_{gi}\dot{q}_i - \dot{v}_{gmi} \tag{22}$$

Additional constraints must be defined to ensure that the control action is lower than the maximum torque allowed for each joint. In this case, the constraint is easy to implement, and the following constraint is included:

$$-\tau_{i,max} \leq \tau_i \leq \tau_{i,max} \tag{23}$$

where $\tau_{i,max}$ is the maximum torque allowed for each joint.

To ensure the contact stability during the docking phase, the tangential forces, $f_t$, must be within the friction cones:

$$f_t < \mu f_n \tag{24}$$

where $f_n$ is the normal component and $\mu$ is the static friction coefficient that characterizes the coupling between the arm end-effector and the target surface. However, constraint (24) must be modified as the optimizer only works with constraints defined by hyperplanes. In this paper, an approximation of the friction cone that consists of inscribing a regular polygon inside the circumference of friction is considered. A coordinate frame centered in the docking point, $j$, is considered which is composed by three vectors perpendicular between them: $a_j$, $b_j$ and $n_j$, where $n_j$ is a unit vector that is normal to the docking plane, and $a_j$, $b_j$ are two vectors within the plane defined by the contact surface. In this case, a new friction coefficient is defined to align the pyramid edges with the real friction cone:

$$\bar{\mu} < \frac{1}{\sqrt{2}}\mu \tag{25}$$

Therefore, the inscribed pyramid is defined by:

$$|a_j f_j| < \bar{\mu}(f_j n_j) \tag{26}$$

$$|b_j f_j| < \bar{\mu}(f_j n_j) \tag{27}$$

where $f_i$ are the contact forces at a given docking point $j$. Equation (26) can be defined by the following inequalities:

$$a_i f_i < \bar{\mu}(f_i n_i) \rightarrow f_i(a_i - \bar{\mu}n_i) < 0 \tag{28}$$

$$-a_i f_i < \bar{\mu}(f_i n_i) \rightarrow -f_i(a_i + \bar{\mu}n_i) < 0 \tag{29}$$

Similar relationships can be obtained from Equation (27):

$$b_i(a_i - \bar{\mu}n_i) < 0 \tag{30}$$

$$-b_i(a_i + \bar{\mu}n_i) < 0 \tag{31}$$

V. RESULTS

This section describes simulation experiments to show the behavior of the trajectory optimization and control approaches presented in the paper. To do this, the on-orbit servicing setup shown in Fig. 2 is considered. Table 2 and Table 3 describe the main robot kinematic dynamic properties considered in the simulations. This robot has $\zeta = 4$ arms with $n = 6$ degrees of freedom each. Each arm has a docking system at their end-effectors. The main objective of the trajectory optimization approach is to determine the body and arms end-effectors' trajectories and interaction forces by using the docking devices located in the workspace.

TABLE II. MAIN DYNAMIC PARAMETERS OF THE ROBOT BODY

|  | Mass (kg) | Height (m) | Inertia (kg·m2) | | | | | |
| --- | --- | --- | --- | --- | --- | --- | --- | --- |
|  |  |  | Ixx | Iyy | Izz | Ixy | Ixz | Iyz |
| **Body Parameters** | 40 | 0.843 | 18.6 | 15.4 | 4.1 | -0.008 | -0.027 | 0.058 |

TABLE III. MAIN DYNAMIC PARAMETERS OF THE ROBOT ARMS

|  | Mass (kg) | Length (m) | Inertia (kg·m2) | | | | | |
| --- | --- | --- | --- | --- | --- | --- | --- | --- |
|  |  |  | Ixx | Iyy | Izz | Ixy | Ixz | Iyz |
| Link 1 | 5.369 | 0.18 | 0.0341 | 0.0353 | 0.0216 | 0 | -0.0043 | -0.0001 |
| Link 2 | 10 | 0.61 | 0.0281 | 0.7707 | 0.7694 | 0.0001 | -0.0156 | 0 |
| Link 3 | 3.9 | 0.57 | 0.0101 | 0.3093 | 0.3065 | 0.0001 | 0.0092 | 0 |
| Link 4 | 2.1 | 0.17 | 0.003 | 0.002 | 0.0026 | 0 | 0 | -0.0002 |
| Link 5 | 1.5 | 0.12 | 0.003 | 0.002 | 0.0026 | 0 | 0 | -0.0002 |
| Link 6 | 0.6 | 0.11 | 0 | 0.0004 | 0.0003 | 0 | 0 | 0 |

## A. Trajectory optimization

In this experiment, the robot body must perform a displacement of 2.1 m along the *x* direction (see Fig. 2). The attitude and the displacement in the other directions should remain constant. Two different simulations are presented in this section. The first one considers that all the target spacecraft surface is able to perform the docking. Therefore, the only constraint for the feasible contact points is that these points should be on the target spacecraft surface. The second simulation includes an additional constraint for the docking positions. These docking positions are located at the corners of a grid with each side measuring 0.5 m (as shown in Fig. 2). Therefore, in this last case, the docking points generated by the trajectory optimization method includes the constraint, $\boldsymbol{p}_{ic}(t \in \mathcal{C}_i) \in \varphi$, where $\varphi$ is the set of feasible docking positions.

Fig. 3 shows the robot body path planning during the manoeuvre generated by the trajectory optimization method (end-effector positions only constrained by the target surface). The *x* component, the *y* component and the *z* component are coloured in blue, red and orange in all the result figures, respectively. As it can be seen in Fig. 3, the displacement of 2.1 m for the robot body in *x* direction is achieved (the displacement in *y* and *z* directions remains constant). However, the robot body orientation doesn't change. Fig. 3 also shows the linear and angular acceleration of the robot body during the trajectory. Fig. 4 represents the interaction forces and end-effector trajectories. Four steps are generated by the trajectory optimization to guide the robot toward the desired location. All the contact positions are within the target surface. Such docking positions are held for the time intervals necessary to let the robot's body move ahead on its path. Fig. 5 shows the interaction forces and arm end-effector trajectories for the same task but considering the docking positions located at the corners of a grid with each side measuring 0.5 m. Several of the available docking devices are used and coordinated between the arms to achieve the desired location for the robot body (see Fig. 6). The trajectory optimization approach is able to generate, not only the trajectories for the robot body and arms, but also the required interaction forces. These forces at the manipulators end-effectors push the robot toward the desired target, taking into account the robot specific dynamics and kinematics.

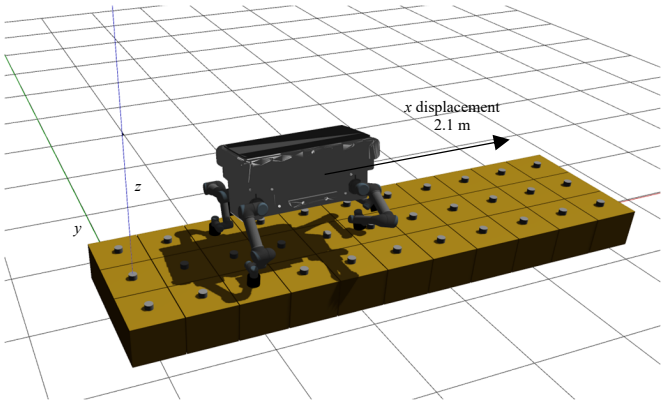

Fig. 2. Simulation setup

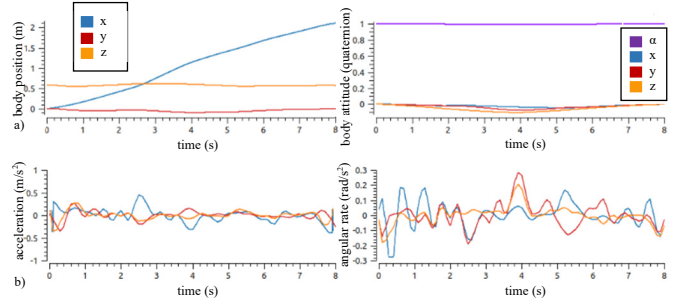

Fig. 3. Robot path planning. a) 3D Position and attitude in quaternions of the robot body. b) Linear and angular acceleration of the robot body. (without end-effector constraints).

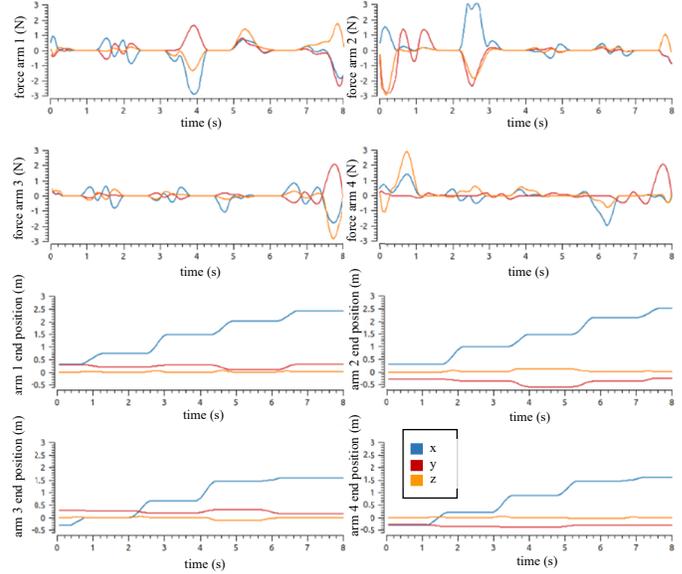

Fig. 4. Multipod path planning. Arms end-effector forces and arms end-effector 3D position. (without end-effector constraints).

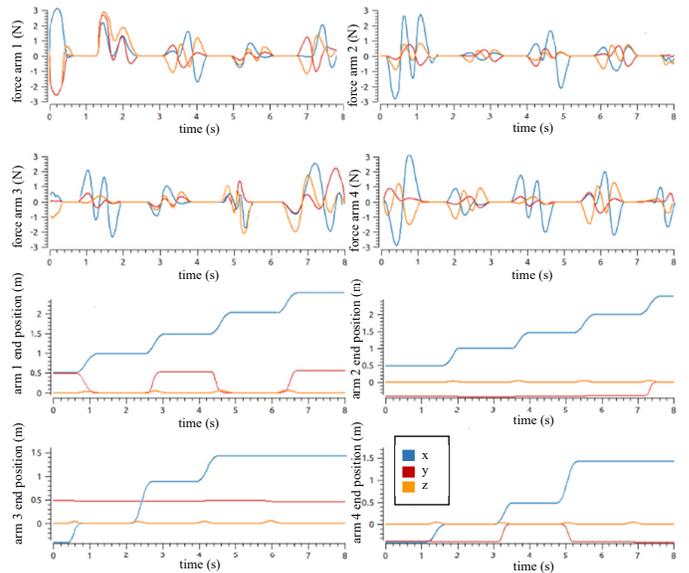

Fig. 5 Robot path planning. Arms end-effector forces and arms end-effector 3D position. (with end-effector constraints).

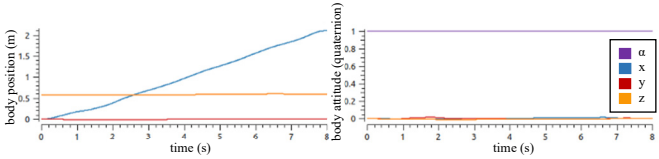

Fig. 6. 3D Position and attitude in quaternions of the robot body. (with end-effector constraints).

## B. Robot control

This section shows the performance of the weighted controller proposed in Section 4. This controller generates the control actions required for the tracking of the trajectories generated by the trajectory optimization problem. As an example, in Fig. 7.a, the joint torque obtained for the tracking of the trajectory of arm 1 is represented (see Fig. 4). The torques remain low during the tracking allowing the correct tracking of the desired trajectory as shown in Fig. 7.b. This last figure represents the control error during the tracking (position and orientation).

Table 4 shows the mean error obtained during the tracking when different controllers are applied. The task space control approaches for complex on-orbit high degrees of freedom robots described in [16] are evaluated. Additionally, a classical PD controller and the optimal controller defined in [9] are also evaluated. It is worth noting that the best behaviour is obtained by using the optimal controller and the approach presented in this paper. However, additional advantages are obtained with the proposed approach. On the one hand, the proposed controller can perform the tracking of both the arms and body trajectories. Both tasks are performed simultaneously and, therefore, the robot body trajectory is also tracked with the same mean error considering the robot free-floating behaviour. On the other hand, the proposed approach allows to define the required constraints to obtain realistic motions considering the on-orbit scenario. The free-floating dynamics of the multi-arm system, and orbital perturbations are considered. In addition, the proposed controller takes into account the required interaction end-effector forces to generate the joint torque control actions.

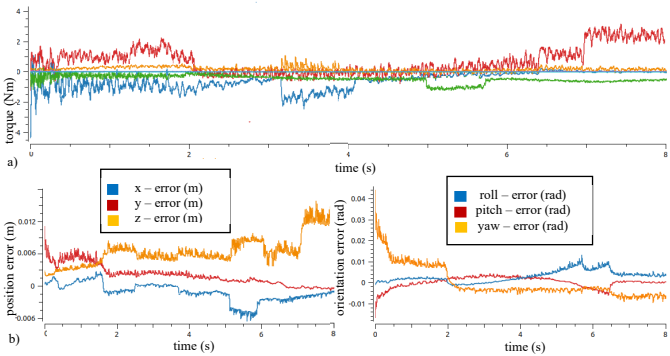

Fig. 7. Controller behaviour for the arm 1. a) Interaction forces. b) Position and orientation errors during the tracking

TABLE IV. MEAN CONTROL ERRORS DURING THE TRACKING

| Controller | Mean error (m) |
|---|---|
| PD Controller | $2.7E-2$ |
| Velocity-based controller | $2.3E-2$ |
| Acceleration controller with inertia matrix pre-multiplication | $1.2E-2$ |
| Acceleration controller without inertia matrix pre-multiplication | $1.3E-2$ |
| Force-based controller | $1.4E-2$ |
| Optimal control approach | $9.3E-3$ |
| Proposed controller | $9.5E-3$ |

## VI. CONCLUSION

A trajectory optimization strategy for motion planning of an on-orbit four-arm robot was presented in this paper. The proposed algorithm generates optimal sequences to guide the arms' movements for coordinated pushing of the contact points (docking), enabling the robot to move both forward and backward. This approach also considers a 3D map obtained by a camera to guarantee the docking points are correctly generated. Consequently, a robust strategy was formulated and tested within a simulated environment, focusing on planning and determining the optimal docking points along with coordinating the articulated movements of the arms. Two different simulations are presented in the paper. The first one considers that all the target spacecraft surface is able to perform the docking. Therefore, the only constraint for the feasible docking points is that these points should be on the target spacecraft surface. The second simulation considers that the docking positions are in the corners of a grid with a side of 0.5 m. In both cases, the proposed approach finds the optimal trajectories and interaction forces, coordinating the docking and non-docking phases and obtaining realistic motions.

In addition, a weighted controller defined as a convex optimization problem is proposed for the tracking of the planned trajectories. This approach is defined in a generic way by optimizing an objective function that allows the system to simultaneously perform a set of tasks. As an example, two tasks are considered: the tracking of the joint trajectories of the arms and the tracking of the body trajectories. As the results show, the control strategy can be applied to track the trajectories generated by the proposed trajectory optimization method, allowing the joint guidance of the robot body and arms.

REFERENCES

[1] M. A. Shoemaker, M. Vavrina, D. E Gaylor, R. Mcintosh, M. Volle, J. Jacobsohn, "OSAM-1 Decommissioning Orbit Design", in *AAS/AIAA Astrodynamics Specialist Conference*, 2020.
[2] B. R. Sullivan, J. Parrish, G. Roesler, "Upgrading In-service Spacecraft with On-orbit Attachable Capabilities", in *2018 AIAA SPACE and Astronautics Forum and Exposition*, 2018.
[3] J. Thangavelautham, A. Chandra, E. Jensen, "Autonomous Robot Teams for Lunar Mining Base Construction and Operation" in *2020 IEEE Aerospace Conference*, 2020, pp 1-16.

[4] H. Mishra, M. De Stefano, C. Ott, "Dynamics and Control of a Reconfigurable Multi-Arm Robot for In-Orbit Assembly", in *10th Vienna International Conference on Mathematical Modelling*, vol. 55, no. 20, 2022, pp. 235-240.

[5] M. Mote, M. Egerstedt, E. Feron, A. Bylard, M. Pavone, "Collision-Inclusive Trajectory Optimization for Free-Flying Spacecraft", *Journal of Guidance, Control, and Dynamics*, vol. 43, 2020, pp. 1–12.

[6] G. Misra, X. Bai, "Task-Constrained Trajectory Planning of Free-Floating Space-Robotic Systems Using Convex Optimization", *Journal of Guidance, Control, and Dynamics*, vol. 40, no. 11, 2017, pp. 2857–3870.

[7] F. Aghili, "Optimal Trajectories and Robot Control for Detumbling a Non-Cooperative Satellite", *Journal of Guidance, Control, and Dynamics*, vol. 43, no. 5, 2020, pp. 981–988.

[8] F. L. Basmadji, K. Seweryn, J.Z. Sasiadek, "Space robot motion planning in the presence of nonconserved linear and angular momenta", *Multibody Syst Dyn*, vol. 50, 2020, pp. 71–96.

[9] J. L. Ramón, R. Calvo, A. Trujillo, J. Pomares, L. Felicetti, "Trajectory Optimization and Control of a Free-Floating Two-Arm Humanoid Robot", *Journal of Guidance, Control, and Dynamics*, vol. 45, no. 9, 2022, pp. 1661-1675.

[10] B. Aceituno-Cabezas et al., "Simultaneous Contact, Gait, and Motion Planning for Robust Multiarmged Locomotion via Mixed-Integer Convex Optimization," in *IEEE Robotics and Automation Letters*, vol. 3, no. 3, pp. 2531-2538, July 2018,

[11] O. Cebe, C. Tiseo, G. Xin, H. -c. Lin, J. Smith and M. Mistry, "Online Dynamic Trajectory Optimization and Control for a Quadruped Robot," in *2021 IEEE International Conference on Robotics and Automation* (ICRA), Xi'an, China, 2021, pp. 12773-12779.

[12] S. Zhou, S. Liu, Z. Lin, Z. Niu, Z. Pan and R. Wang, "Cascade Trajectory Optimization with Phase Duration Adaption and Control for Wheel-Legged Robots Overcoming High Obstacles," 2023 *International Conference on Advanced Robotics and Mechatronics* (ICARM), Sanya, China, 2023, pp. 832-839,

[13] D. Pinard, S. Reynaud, P. Delpy, S. E. Strandmoe, "Accurate and autonomous navigation for the ATV", *Aerospace Science and Technology*, vol. 11, no. 6, 2007, pp. 490-498.

[14] J. Pomares, L. Felicetti, G. J. García, J. L. Ramón. "Spacecraft Formation Keeping and Reconfiguration Using Optimal Visual Servoing". *J Astronaut Sci* vol. 71, no. 19, 2024.

[15] J. L. Ramón, J. Pomares, L. Felicetti, "Direct visual servoing and interaction control for a two-arms on-orbit servicing spacecraft", *Acta Astronautica*, vol. 192, 2022, pp. 368-378.

[16] J. L. Ramón, J. Pomares, L. Felicetti, "Task space control for on-orbit space robotics using a new ROS-based framework", *Simulation Modelling Practice and Theory*, vol. 127, 2023, 102790.